\documentclass{article}
\usepackage[T1]{fontenc}
\usepackage[utf8]{inputenc}
\usepackage[normalem]{ulem}
\usepackage{microtype}
\usepackage{graphicx}
\usepackage{booktabs}
\usepackage{hyperref}
\usepackage{amsmath}
\usepackage{amssymb}
\usepackage{mathtools}
\usepackage{amsthm}
\usepackage[capitalize,noabbrev]{cleveref}

\usepackage[accepted]{icml2026}
\makeatletter
\renewcommand{\ICML@appearing}{\textit{Mechanistic Interpretability Workshop at the
$\mathit{43}^{rd}$ International Conference on Machine Learning},
Seoul, South Korea, 2026.
Copyright 2026 by the author(s).}
\makeatother
\icmltitlerunning{Residue-Level Attributions in Protein Language Models Do Not Recover Allergen Epitopes}

\begin{document}

\twocolumn[
  \icmltitle{Residue-Level Attributions in Protein Language Models Do Not Recover Allergen Epitopes}
  


  
\icmlsetsymbol{equal}{*}

\begin{icmlauthorlist}
\icmlauthor{Jianzhou Yao}{siaf,eth}
\icmlauthor{Anxiong Song}{siaf,eth}
\icmlauthor{Katja Baerenfaller}{siaf,sib}
\icmlauthor{Damir Zhakparov}{siaf,sib}
\end{icmlauthorlist}

\icmlaffiliation{siaf}{Swiss Institute of Allergy and Asthma Research, Davos, Switzerland}
\icmlaffiliation{sib}{Swiss Institute of Bioinformatics, Lausanne, Switzerland}
\icmlaffiliation{eth}{ETH Zurich, Zurich, Switzerland}

\icmlcorrespondingauthor{Jianzhou Yao}{yaojia@ethz.ch}

\icmlkeywords{Trustworthy AI; protein language models; mechanistic interpretability; allergenicity prediction; epitopes; explainable AI}

\vskip 0.3in
]



\printAffiliationsAndNotice{}  

\begin{abstract}
Deep allergenicity classifiers are increasingly used in safety screening of novel foods, and recent protein language models have substantially
improved protein-level allergenicity prediction. However, whether their explanations
capture biologically meaningful information remains unclear.
We introduce an epitope-grounded residue-level benchmark for
quantitatively evaluating attribution faithfulness in protein allergenicity
models. Across frozen ESM-2, multi-task ESM-2, and DeepPlantAllergy,
protein-level classification was robust, yet classification-head explanation signals did not significantly exceed random in their
residue-level alignment with annotated epitopes across AUROC, AUPRC, and
Precision@$k$.
Integrated Gradients identified residues that were functionally important to
the model, but not overlapping annotated epitopes. Saturation
mutagenesis further suggested classifiers may rely on physicochemical and
compositional sequence features rather than epitope-specific mechanisms.
Residue-level importance signals should therefore not be interpreted as
immunological explanations for safety screening or hypoallergen design without quantitative validation. Code available \href{https://github.com/Jeffateth/XAllergen2.0-paper}{\texttt{here}}.
\end{abstract}

\section{Introduction}
\label{sec:intro}

Allergenicity predictors are increasingly used in safety screening of novel
foods and recombinant proteins, complementing experimental assays and expert reviews~\cite{fernandez_novel_food_2021,mullins_efsa_2022}. Recent advances in protein language models have substantially improved computational protein
annotation, including allergenicity prediction~\cite{lin_esm2_2023,he_deepalgpro_2023,dhouib_deepplantallergy_2025}. However, predictive accuracy alone is insufficient if models rely on immunologically implausible features.


In allergy, immune recognition targets specific regions of an allergenic protein called epitopes rather than the full protein, with epitope-level recognition shaping sensitization, clinical reactivity, and cross-reactivity~\cite{moiniche_epitope_mapping_2026}. We therefore ask: \emph{do strong protein-level allergenicity classifiers produce residue-level explanations that align with experimentally annotated epitopes?}

Recent allergenicity classifiers often visualize attribution scores, such as
Integrated Gradients (IG)~\cite{sundararajan_ig_2017}, as residue-level heatmaps and show qualitative agreement with epitope motifs on individual protein examples~\cite{he_deepalgpro_2023,dhouib_deepplantallergy_2025,liu_allergenai_2025}, but their quantitative epitope alignment remains unknown.

We distinguish \textbf{model faithfulness} -  whether an explanation
highlights residues that influence the model's output, from \textbf{immunological
faithfulness} - whether those residues align with experimentally observed
immune-recognition sites. 

Quantitative ground-truth evaluation of attributions has been pursued in other domains, including pixel-level masks for visual question answering~\cite{arras_clevrxai_2022} and clinician-validated features in medical imaging~\cite{makino_human_machine_2022,arun_saliency_2021}; curated epitope databases offer an analogous source of ground truth for the immunological faithfulness of allergenicity classifiers.

We therefore adopted rank-based metrics (AUROC, AUPRC) established for evaluating supervised epitope
predictor performance under strong class imbalance~\cite{clifford_bepipred3_2022,hoie_discotope3_2024,israeli_caliber_2024,zeng_graphbepi_2023,cia_critical_2023},
applied here to explanation signals from allergenicity classifiers rather than
models trained directly for epitope prediction. We further include a multi-task learning (MTL) model with auxiliary residue-level epitope supervision as an interventional diagnostic of epitope-relevant feature use.

Our contributions are:

\begin{figure*}[t]
  \centering
  \includegraphics[width=\textwidth]{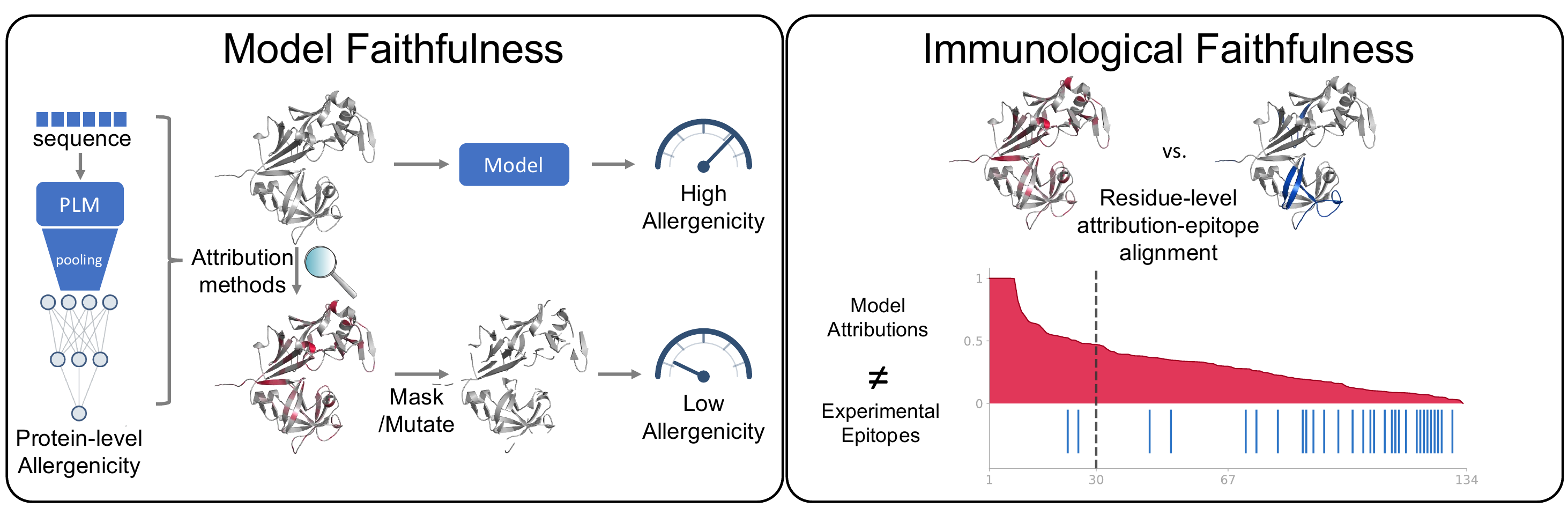}
  \caption{
Graphical illustration of the two faithfulness criteria distinguished in this work.
Left: Model faithfulness: residues identified by attribution methods causally influence the allergenicity prediction when masked, illustrated using the 3D structure of the allergenic protein P54958.
Right: Immunological faithfulness: these residues need not coincide with experimentally annotated epitopes; blue ticks, IEDB epitopes; red profile, Integrated Gradients scores from the allergenic protein K7AKJ8 ranked from highest to lowest; dashed line, Precision@k cutoff.
  }
  \label{fig:faithfulness_concept}
\end{figure*}

\begin{enumerate}
  \item We provide an epitope-grounded benchmark for quantitatively evaluating residue-level
        immunological faithfulness of allergenicity model
        explanations, using AUROC, AUPRC, and Precision@$k$.

  \item We show that strong allergenicity classifiers consistently fail to produce
        epitope-aligned attribution signals better than random across all models and metrics.

\item We demonstrate a dissociation between model and immunological faithfulness: IG is model-faithful under masking, yet saturation mutagenesis reveals that models are sensitive to global physicochemical rather than epitope-specific features.
\end{enumerate}

\section{Benchmark and Methods}
\label{sec:methods}

\paragraph{Residue-level faithfulness definition.}
Let $x=(x_1,\ldots,x_L)$ be a protein sequence and $F(x)\in[0,1]$ a
protein-level allergenicity classifier. A residue-level scoring method
produces $s=(s_1,\ldots,s_L)$, where larger $s_i$ indicates greater
importance of residue $i$. For allergenic proteins with experimentally
annotated allergy-associated MHC class II epitopes, we define a binary mask
$y=(y_1,\ldots,y_L)$, where $y_i=1$ if residue $i$ belongs to an annotated
epitope region after mapping to the parent protein. Immunological
faithfulness asks whether high-scoring residues under $s$ correspond to
positive residues under $y$; model faithfulness asks whether they
causally influence $F(x)$.

\paragraph{Protein-level allergenicity data.}
Protein-level classifiers were trained and evaluated using the released
DeepAlgPro train/test FASTA files~\cite{he_deepalgpro_2023}. Exact sequence
overlaps with the epitope benchmark were removed prior to training and
evaluation. The resulting cleaned held-out test set contained 1,377 sequences.

\paragraph{Epitope-grounded benchmark curation.}
Because allergenicity is mediated by epitope recognition, including
conformational or linear B-cell epitopes and linear peptide fragments
presented to T cells by antigen-presenting cells, we curated experimentally
validated allergy-associated epitopes from the Immune Epitope Database
(IEDB)~\cite{vita_iedb_2019}. We retained positive assays, restricted to
allergic-disease contexts, and included both B-cell and T-cell epitopes under
MHC class II restriction. Full-length parent proteins were retrieved through
UniProt~\cite{uniprot_2023} and NCBI~\cite{sayers_ncbi_2025}. We excluded
sequences with non-canonical amino acids, invalid epitope coordinates, viral
or human origin, individual epitopes covering more than 25\% of the parent
protein, merged epitope coverage exceeding 75\%, duplicate sequences, and
sequences exceeding the ESM-2 context length.

Candidate negatives were retrieved from UniProtKB/Swiss-Prot reviewed
entries, excluding viral and human proteins, allergen annotations,
allergenic-related text, antigen- or cancer-related entries, and
positive-homologous sequences. We further required evidence at protein level,
annotation score 5, and annotated three-dimensional structural availability.
Homology-aware splitting prevented clustered sequences from appearing in both
training and evaluation splits. Full curation details are provided in
Appendix~\ref{app:data}.

\paragraph{Models.}
We evaluated three classifiers. 

\textbf{Frozen ESM-2} used
\texttt{esm2\_t6\_8M\_UR50D} with learned attention pooling and a two-layer
MLP classification head~\cite{lin_esm2_2023}. The ESM-2 backbone was frozen;
only the attention pooling and classifier layers were trained.

\textbf{MTL ESM-2} shares the same frozen ESM-2 backbone, initialized from
the trained classification head best checkpoint, and adds an auxiliary residue-level epitope prediction head. The model was trained jointly
on protein-level allergenicity labels and residue-level epitope masks:
\begin{equation}
  \mathcal{L}
  =
  \lambda_{\mathrm{cls}}\mathcal{L}_{\mathrm{cls}}
  +
  \lambda_{\mathrm{epi}}\mathcal{L}_{\mathrm{epi}}.
\end{equation}
MTL ESM-2 served as a principled intervention motivated by the MTL principle
that auxiliary supervision on related tasks can improve primary task
generalization through shared
representation~\cite{caruana_multitask_1997,ruder_mtl_overview_2017}.
If epitope-relevant residue information were accessible to the allergenicity
classifier, injecting residue-level epitope supervision should improve
protein-level classification or produce more epitope-aligned
classification-head explanations. The auxiliary residue head was a diagnostic
supervised signal, not a proposed epitope predictor or classification head explanation.

\textbf{DeepPlantAllergy} was included as a published external allergenicity architecture~\cite{dhouib_deepplantallergy_2025}, retrained from scratch on our curated dataset to enable controlled comparison and avoid data leakage. Architecture details are in Appendix~\ref{app:implementation}.

\paragraph{Residue-level scores.}
We evaluated IG, occlusion, and the auxiliary MTL
residue head. IG computes path-integrated sensitivity from a
zero-embedding baseline to the input embedding~\cite{sundararajan_ig_2017}.
Occlusion masked one residue at a time with the ESM-2 mask token and scored the
resulting allergenicity decrease, $s_i^{\mathrm{occ}}=F(x)-F(x_{\setminus i})$.
Additional signals, including pooling attention weights, Gradient$\times$Input,
SmoothGrad-IG, and MTL classifier attributions, are defined and evaluated in
Appendix~\ref{app:signal_definitions} and~\ref{app:all_signals}.

\paragraph{Metrics.}
Residue-level evaluation was performed on the 61 held-out splitB allergenic
proteins with epitope annotations. Following epitope-prediction evaluations,
we used established metrics AUROC and AUPRC to assess threshold-independent residue-level ranking
under class imbalance~\cite{clifford_bepipred3_2022,hoie_discotope3_2024,israeli_caliber_2024,zeng_graphbepi_2023,cia_critical_2023}.
For each protein, the score vector $s$ was compared with the binary epitope mask
$y$. We computed metrics within each protein and then averaged across proteins. Random baselines were computed identically using repeated random residue scores. AUROC measures whether epitope
residues are ranked above non-epitope residues, while AUPRC emphasizes
performance on sparse positive residues. We additionally reported Precision@$k$ as a top-ranked localization metric,
where $k$ was set for each protein to their respective number of annotated
epitope residues, $k=\sum_i y_i$:
\begin{equation}
\mathrm{P@}k
=
\frac{|\mathrm{Top}_k(s)\cap \{i:y_i=1\}|}{k}.
\end{equation}
For statistical testing, each non-random residue-level signal was compared
against the corresponding random baseline from the same model family using a
paired Wilcoxon signed-rank test across proteins. Tests were performed
separately for AUROC, AUPRC, and Precision@$k$, and Benjamini--Hochberg
correction was applied within each metric family.

\paragraph{Functional validation.}
To test model faithfulness, we ranked residues in the Frozen ESM-2 baseline by
IG and simultaneously masked the top-$k$ fraction, measuring
$\Delta p = F(x) - F(x_{\mathrm{masked}})$.
The control consisted of random masking of equal size; IG-guided masking
should reduce $F(x)$ more if IG is model-faithful.
Full sweep details are in Appendix~\ref{app:masking}.

\paragraph{Saturation mutagenesis.}
To characterize the decision basis of the Frozen ESM-2 baseline classifier,
we used \textit{in silico} saturation mutagenesis, a perturbation-based interpretability
approach in which systematic single-residue substitutions are used to measure
changes in model output~\cite{lim_antibody_binders_2022,koido_transcriptional_2024}. While IG and occlusion identify \emph{which} residues the model relies on, saturation mutagenesis probes \emph{why}: by exhaustively substituting each residue and measuring the resulting change in predicted allergenicity, it tests whether this reliance reflects epitope-specific residue identity or broader physicochemical properties shared across many positions.
Each residue was substituted with each of the other 19 alternative amino acids, recording
$\Delta p_{i,a\to b} = F(x) - F(x_{i:a\to b})$.
Effects were aggregated by original amino acid and by coarse residue classes
(charge, polarity, hydrophobicity, aromaticity); class definitions and
transition summaries are in Appendix~\ref{app:mutagenesis_classes}.

\section{Results and Discussion}
\label{sec:results}

\subsection{Strong classification does not imply epitope-relevant attribution}

\begin{figure}[t]
  \centering
  \includegraphics[width=\linewidth]{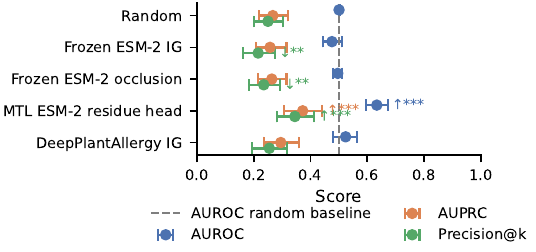}
  \caption{
Residue-level alignment between model-derived scores and IEDB epitope annotations for representative methods. Error bars denote 95\% bootstrap confidence intervals (CIs). Significance markers denote adjusted p-values: $^{*}p<0.05$, $^{**}p<0.01$, $^{***}p<0.001$; arrows indicate direction ($\uparrow$ increase, $\downarrow$ decrease). Numerical values are reported in \cref{tab:alignment_summary}.
}
  \label{fig:residue_alignment}
\end{figure}

We first ask whether strong protein-level performance is accompanied by epitope-aligned attribution signals. Despite all evaluated models achieved strong protein-level allergenicity classification
on the held-out DeepAlgPro test set (\cref{tab:protein_performance}), no model-derived attribution signal exceeded its
corresponding random baseline, and several were significantly lower than random,
indicating that these signals did not capture epitope-relevant information and
may instead reflect other model-internal or biological information
(\cref{fig:residue_alignment}). This pattern persisted in the full signal
comparison (\cref{fig:all_signals_significance}). A label-scrambling negative control confirmed that the evaluation was well
calibrated: scrambling epitope labels substantially reduced the performance
of the supervised MTL residue head, while leaving the already near-random
attribution methods essentially unchanged (\cref{fig:label_scrambling}).

We next test whether explicit epitope supervision improves alignment. The auxiliary residue-level head in MTL ESM-2 learned directly
supervised epitope signals, but incorporating this supervision improved neither
protein-level classification performance nor the alignment between
classification-head attributions and experimentally validated epitope
positions. This null result was consistent with prior work showing that MTL
improves generalization only when tasks cooperate at the representation
level~\cite{standley_taskgrouping_2020,wu_mtl_transfer_2020}. Although a frozen backbone could in principle limit representation-level transfer, we additionally trained an exploratory MTL variant with the top ESM-2 layer unfrozen during joint training; despite enabling representation-level adaptation, it showed the same qualitative pattern of weak classification-head epitope alignment (\cref{fig:all_signals_significance}). The failure of explicit epitope
supervision to transfer to the classification head therefore provided evidence
that protein-level allergenicity classification can be achieved without relying
on localized epitope-relevant features.

\begin{table}[t]
\centering
\caption{Protein-level classification performance on the held-out DeepAlgPro
test set ($n=1377$). Additional metrics in \cref{tab:protein_full}.}
\label{tab:protein_performance}
\footnotesize
\setlength{\tabcolsep}{7pt}
\begin{tabular}{lccc}
\toprule
Model & AUROC & F1 & MCC \\
\midrule
Frozen ESM-2     & 0.970 & 0.917 & 0.835 \\
MTL ESM-2        & 0.962 & 0.910 & 0.821 \\
DeepPlantAllergy & 0.974 & 0.939 & 0.878 \\
\bottomrule
\end{tabular}
\end{table}

\subsection{IG is model-faithful but immunologically misaligned}

Poor epitope alignment could reflect either uninformative attributions or
accurate identification of model-relevant residues that were not epitope
positions. We distinguished these possibilities for the Frozen ESM-2 baseline
using the functional masking test defined in \cref{sec:methods}.
 
\begin{figure}[t]
  \centering
  \includegraphics[width=\linewidth]{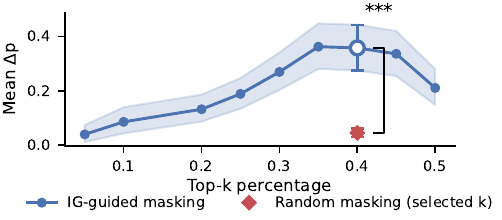}
  \caption{
  Functional masking analysis for Frozen ESM-2 using IG.
  The blue curve shows the mean change in predicted allergenicity when masking the top-$k$ IG-ranked
  residues; shaded regions denote bootstrap
  95\% CIs. The open blue marker denotes the selected $k$.
  The bracket compares IG-guided and random masking at $k=40\%$ using a paired
  two-sided Wilcoxon signed-rank test across proteins; *** indicates
  $p<0.001$. 
  }
  \label{fig:ig_masking}
\end{figure}

Residues were ranked by IGs, and the top-$k$ fraction was simultaneously
masked to assess their influence on the model output. The primary analysis was restricted to high-confidence predictions
($F(x)>0.70$, $n=46$ proteins). At the selected operating point $k=40\%$,
IG-guided masking reduced predicted allergenicity significantly more than
random masking of equal size (\cref{fig:ig_masking}), with the effect persisting without the confidence filter (\cref{fig:ig_masking_unfiltered}).
Details of operating point selection are  provided in Appendix~\ref{app:masking}.

These results indicated that IG identified residues that causally influenced the model’s
predictions under this perturbation test. Yet, given the  weak alignment with annotated
epitopes, this implied the gap between model faithfulness and
immunological faithfulness: the model relied on residue-level signals that are predictive but not epitope-specific.

\subsection{Mutagenesis reveals global physicochemical rather than epitope-specific sensitivity}

To further characterize the decision basis of the Frozen ESM-2 classifier, we
performed saturation mutagenesis on the 37 proteins validated by the IG-masking
criterion at $k=40\%$.

\begin{figure}[t]
  \centering
  \includegraphics[width=\linewidth]{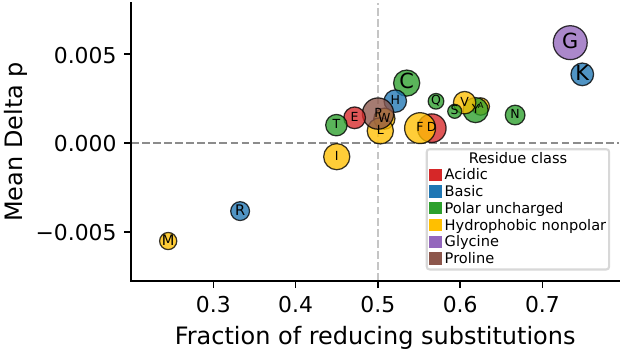}
\caption{
Saturation mutagenesis sensitivity by amino acid. Each point represents a wild-type amino acid; point size reflects its frequency across all evaluated proteins. The $x$-axis shows the fraction of substitutions that reduced predicted allergenicity (i.e., the fraction of substitutions originating from that amino acid with $\Delta p > 0$ across all positions and proteins). Positive values indicate a net reduction in predicted allergenicity.
}
\label{fig:transition_scatter}
\end{figure}

Individual amino acids differed substantially in both the fraction of
substitutions that reduce predicted allergenicity and the mean effect size
$\Delta p$ (\cref{fig:transition_scatter}). Methionine and arginine showed
negative mean $\Delta p$, indicating that substituting them tended, on average,
to increase predicted allergenicity, while glycine exhibited the highest
fraction of reducing substitutions and the largest mean $\Delta p$, followed
by lysine and cysteine.

To assess whether this sensitivity reflects biochemical structure,
the class-level transition heatmaps revealed that it was
structured along physicochemical axes
(residue classes defined in
\cref{tab:residue_classes};
\cref{fig:charge_polarity_heatmap,fig:hydrophobicity_heatmap}).
Charge-altering transitions produced larger effects than within-class
substitutions, consistent with the role of charge state in electrostatic
interactions at protein surfaces and epitope recognition
sites~\cite{zhou_arah2_critical_2024,zhou_arah2_structural_2022}.
Within-aromatic and within-hydrophobic-aliphatic substitutions showed
smaller effects, indicating lower model sensitivity to changes that preserved
hydrophobic or aromatic character; the solvent-accessible surface area of
aromatic residues correlates with IgE-binding capacity in food
allergens~\cite{zhou_arah2_structural_2022}. Glycine and proline showed elevated cross-class effects in both heatmaps:
glycine substitutions introduce a side chain where none existed, altering
backbone turn capacity, while proline removal relieves its constrained
$\phi$ angle and \textit{cis}-peptide bond preference. Cysteine sensitivity
reflects the multifunctional role of its sulfhydryl group, which participates
in disulfide-stabilized structure, enzymatic active sites, and
post-translational modification sites including
lipidation~\cite{pekar_stability_2018}.

Together, these structured sensitivities were consistent with reliance on
global physicochemical and compositional features rather than direct
localization within annotated linear epitope regions. Mutagenesis effects across all 61 splitB proteins
were significantly larger outside epitope residues than inside
(Wilcoxon $p=0.021$; \cref{fig:epitope_vs_non_epitope}), further
arguing against epitope-specific sensitivity.

\section{Conclusion and Future Work}
\label{sec:conclusion_future}

We demonstrated a consistent dissociation between predictive performance,
model faithfulness, and immunological faithfulness: while the three tested models achieved strong protein-level classification performance and produced reliable attribution signals, they did not reliably recover annotated MHC class II epitopes.

The present analysis is limited by incomplete and biased IEDB annotations, restriction
to linear MHC-II-derived residue masks, modest benchmark size, and the use of
\textit{in silico} perturbations that approximate rather than experimentally
validate immune recognition.

These findings have direct implications for downstream use. In hypoallergen
design, modifying residues highlighted by attribution methods may not reduce
epitope content if these signals are not epitope-aligned. In safety screening,
attribution visualizations may support model auditing but should not be
interpreted as evidence of epitope-level understanding. 
More broadly, these results motivate biologically constrained predictors that incorporate
structural and immunological priors, and call for standardized,
epitope-grounded benchmarks to evaluate whether model explanations capture
biologically meaningful mechanisms. They also motivate future mechanistic interpretability studies using probing and causal interventions to investigate how allergenicity-relevant information is represented within protein language models.
\bibliographystyle{icml2026}
\bibliography{reference}

\clearpage
\appendix

\section{Dataset Curation Details}
\label{app:data}

IEDB epitope entries are compiled from two primary sources: peer-reviewed
literature curated by IEDB staff and direct submissions from
researchers~\cite{fleri_iedb_curation_2017}.

\paragraph{Positive epitope set.}
The initial IEDB-derived positive table contained 4,221 epitope rows after
accession resolution and sequence retrieval; they consisted primarily of linear
epitopes, as no conformational epitopes were available under these filters.
Removing non-canonical amino acid sequences left 4,146 rows. Proteins
requiring coordinate clipping were removed rather than clipped, leaving 3,991
rows. Individual epitopes spanning more than 25\% of the parent protein were
removed as biologically implausible for linear immune-recognition fragments.
Surviving intervals were merged, and proteins whose merged epitope coverage
exceeded 75\% were removed as likely annotation artefacts. This yielded 416
proteins after the coverage filter, 391 after viral and human-source removal,
and 374 after the maximum-length filter ($\leq 1022$ aa, imposed by the ESM-2
context window).

\paragraph{Negative set.}
The initial UniProt-derived negative set contained 11,139 reviewed entries after
applying the filters described in \cref{sec:methods}. Removing non-canonical
sequences left 11,118 entries. Applying the maximum-length filter
($\leq 1022$ aa) left 10,140 entries. Sequence-level redundancy was removed
with \texttt{mmseqs easy-cluster} ($\geq$40\% sequence identity, $\geq$80\%
bidirectional coverage, \texttt{--cov-mode 0}), yielding 7,572 cluster
representatives. Positive-homologous negatives were identified with
\texttt{mmseqs easy-search} against the full positive FASTA ($\geq$30\%
identity, $\geq$80\% coverage, \texttt{--cov-mode 0}) and removed, leaving
7,477 clean representatives. Nearest-length greedy matching (random state 13) subsampled the 7,477 clean
negative representatives to 374 negatives, matching the 374 positive proteins
and producing a 1:1 positive-negative benchmark pool before splitting.

\paragraph{Splits.}
Positive proteins were grouped by their MMseqs2 cluster representative; negative
proteins, already reduced to one representative per cluster, each formed their
own group. A custom greedy algorithm assigned entire groups to the 80/20 split
to jointly balance total samples, positives, and negatives across splits
(random state 13), ensuring no homologous positive sequences appeared in both
splits. This yielded 313 positive and 299 negative proteins in splitA, and 61
positive and 75 negative proteins in splitB. The residue-level faithfulness
benchmark used the 61 splitB positive proteins with epitope masks. The mean
positive-test epitope density was approximately 0.250.

\begin{figure}[h]
  \centering
  \includegraphics[width=\linewidth]{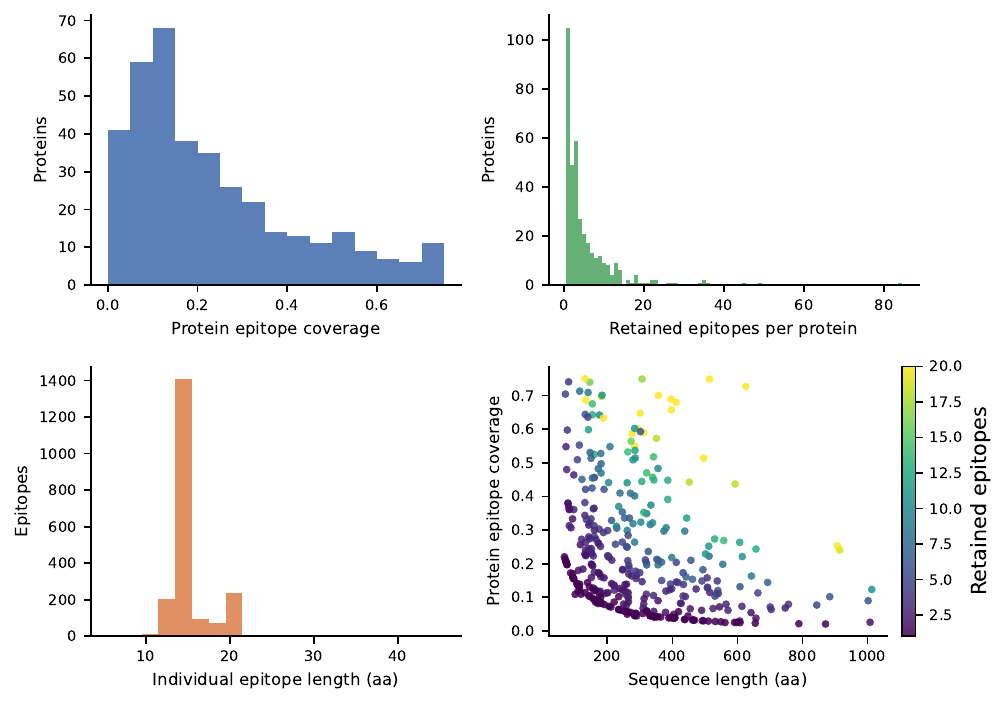}
  \caption{
  Positive epitope benchmark profile. The panels summarize protein-level
  epitope coverage, retained epitopes per protein, individual epitope lengths,
  and the relationship between sequence length, coverage, and retained epitope
  count.
  }
  \label{fig:positive_dataset_profile}
\end{figure}

\section{Implementation Details}
\label{app:implementation}

\paragraph{Frozen ESM-2.}
The frozen ESM-2 classifier used \texttt{esm2\_t6\_8M\_UR50D}, hidden
dimension 128, dropout 0.3, threshold 0.5, IG with 50 steps, and maximum
sequence length 1022. The ESM-2 backbone was frozen; only learned attention
pooling and the classification head were trained.

\paragraph{MTL ESM-2.}
The MTL model was initialized from the trained Frozen ESM-2 checkpoint. The
ESM-2 backbone, learned attention pooling module, and protein-level
classification head were reused, and a new auxiliary residue-level epitope head
was added on top of the per-residue ESM-2 hidden states. In the reported
frozen-MTL setting, the ESM-2 backbone remained frozen; the trainable components
were the protein-level head, attention-pooling parameters, and the newly
initialized auxiliary residue-level epitope head.

Training used mixed batches containing allergenic proteins with residue-level
epitope masks and non-allergenic proteins without residue-level epitope
supervision. Protein-level binary cross-entropy was computed for all sequences.
Residue-level binary cross-entropy was computed only at valid residue positions
for positive proteins with epitope annotations, using a residue mask to exclude
proteins without epitope supervision from the epitope loss. The residue loss
used positive-class weighting to reduce imbalance between sparse epitope
residues and more abundant non-epitope residues. Specifically, for residue $i$,
\begin{equation}
  \ell_i
  =
  - w_+ y_i \log \sigma(z_i)
  -
  (1-y_i)\log(1-\sigma(z_i)),
\end{equation}
and the masked residue loss is
\begin{equation}
  \mathcal{L}_{\mathrm{epi}}
  =
  \frac{\sum_i m_i \ell_i}{\sum_i m_i},
  \qquad
  w_+ = \frac{N_-}{N_+}.
\end{equation}
Here $z_i$ is the residue-level logit, $m_i$ indicates valid supervised
residue positions, and $w_+$ is computed from the ratio of non-epitope to
epitope residues in the positive training proteins. This weighting avoided
downsampling the majority non-epitope class, which is important given the
limited amount of experimental epitope supervision. Protein-level class
weighting was disabled because the cleaned DeepAlgPro training split was close
to balanced.

Hyperparameters: batch size 24, maximum 30 epochs, early-stopping patience 5,
learning rate $10^{-3}$, weight decay $10^{-4}$, $\lambda_{\mathrm{cls}}=1.0$,
$\lambda_{\mathrm{epi}}=0.5$, epitope-head hidden dimension 128, validation
fraction 0.1, 100 random attribution draws, and IG internal batch size 1. The
best checkpoint was selected by validation total loss.

\paragraph{DeepPlantAllergy.}
The DeepPlantAllergy benchmark used ESM-1b embeddings with dimension 1280,
convolutional filters, a three-layer bidirectional LSTM, multi-head
self-attention with eight heads, adaptive average pooling, and fully connected
layers. The maximum sequence length is 1000 for the ESM-1b embedding workflow.

\paragraph{Exploratory top-1-unfrozen MTL.}
As a robustness control for the frozen-backbone MTL setting, we additionally
trained an exploratory MTL checkpoint in which the top ESM-2 layer is unfrozen
during joint training. This checkpoint was included in the supplementary
all-signal analysis but not in the primary protein-level model comparison.

\section{Residue-Level Signal Definitions}
\label{app:signal_definitions}

The main paper reports a curated subset of residue-level scores. Here we
define the full active signal inventory.

\paragraph{Integrated Gradients.}
Integrated Gradients attributes importance by integrating gradients along a
path from a baseline embedding $e'$ to the observed input embedding
$e$~\cite{sundararajan_ig_2017}:
\begin{equation}
s_i^{\mathrm{IG}}
=
\left\|
(e_i-e'_i)
\int_{\alpha=0}^{1}
\nabla_{e_i}F(e' + \alpha(e-e'))\,d\alpha
\right\|_1.
\end{equation}
The integral was approximated with a finite number of interpolation steps.
Path-based attributions can accumulate noisy gradients along the integration
trajectory, motivating comparison with additional attribution
variants~\cite{kapishnikov_guided_2021}.

\paragraph{Gradient$\times$Input.}
For residue embedding $e_i$, Gradient$\times$Input is:
\begin{equation}
s_i^{\mathrm{GxI}}
=
\left\|
e_i \odot \nabla_{e_i}F(x)
\right\|_1.
\end{equation}
This provides a local first-order sensitivity estimate~\cite{ancona_gradient_2018}.

\paragraph{SmoothGrad-IG.}
SmoothGrad-IG averages IG scores over noisy embedding perturbations:
\begin{equation}
s_i^{\mathrm{SG\text{-}IG}}
=
\frac{1}{M}
\sum_{m=1}^{M}
s_i^{\mathrm{IG}}(e+\epsilon^{(m)}),
\quad
\epsilon^{(m)}\sim \mathcal{N}(0,\sigma^2 I).
\end{equation}
This tested whether epitope alignment improved when attribution scores were
stabilized by noise averaging~\cite{smilkov_smoothgrad_2017}.

\paragraph{Occlusion.}
Occlusion masked one residue at a time and measured the decrease in predicted
allergenicity:
\begin{equation}
  s_i^{\mathrm{occ}} = F(x) - F(x_{\setminus i}).
\end{equation}

\paragraph{Attention weights.}
For models with learned attention pooling, the normalized pooling weights are
used as residue-level scores. We treated these as model-internal signals rather
than guaranteed explanations, since attention weights are not necessarily
faithful causal explanations of model predictions.

\paragraph{Auxiliary MTL residue head.}
The auxiliary MTL residue head directly predicts a residue-level epitope
probability for each position. It was reported as a diagnostic supervised signal
indicating whether the injected epitope labels are learnable by an auxiliary
head, not as a post-hoc explanation of the protein-level classification head.

\paragraph{Random baseline.}
The random baseline assigned random scores to residues within each protein,
recomputed over repeated draws before aggregation. This controlled for
protein-specific epitope density and metric behavior.

\section{Additional Protein-Level Classification Metrics}
\label{app:protein_full}

\begin{table}[h]
\centering
\caption{Full protein-level classification metrics on the held-out DeepAlgPro
test set ($n=1377$).}
\label{tab:protein_full}
\scriptsize
\setlength{\tabcolsep}{3.0pt}
\resizebox{\columnwidth}{!}{%
\begin{tabular}{lcccccc}
\toprule
Model & AUROC & Prec. & Rec. & F1 & MCC & Acc. \\
\midrule
Frozen ESM-2     & 0.970 & 0.901 & 0.933 & 0.917 & 0.835 & 0.917 \\
MTL ESM-2        & 0.962 & 0.886 & 0.936 & 0.910 & 0.821 & 0.910 \\
DeepPlantAllergy & 0.974 & 0.939 & 0.939 & 0.939 & 0.878 & 0.939 \\
\bottomrule
\end{tabular}%
}
\end{table}

\section{All-Signals' Residue-Level Faithfulness}
\label{app:all_signals}

The main paper reports a curated subset of residue-level signals to preserve
readability. Here we report the full active signal inventory across all
available model families. For Frozen ESM-2 and DeepPlantAllergy, this includes
attention weights, Integrated Gradients, Gradient$\times$Input, SmoothGrad-IG,
occlusion, and random baselines. For MTL ESM-2, the same classification-head
signals are reported together with the auxiliary residue head. Exploratory
supplementary checkpoints are included only when their probe rows are available.

\begin{figure}[h]
  \centering
  \includegraphics[width=\linewidth]{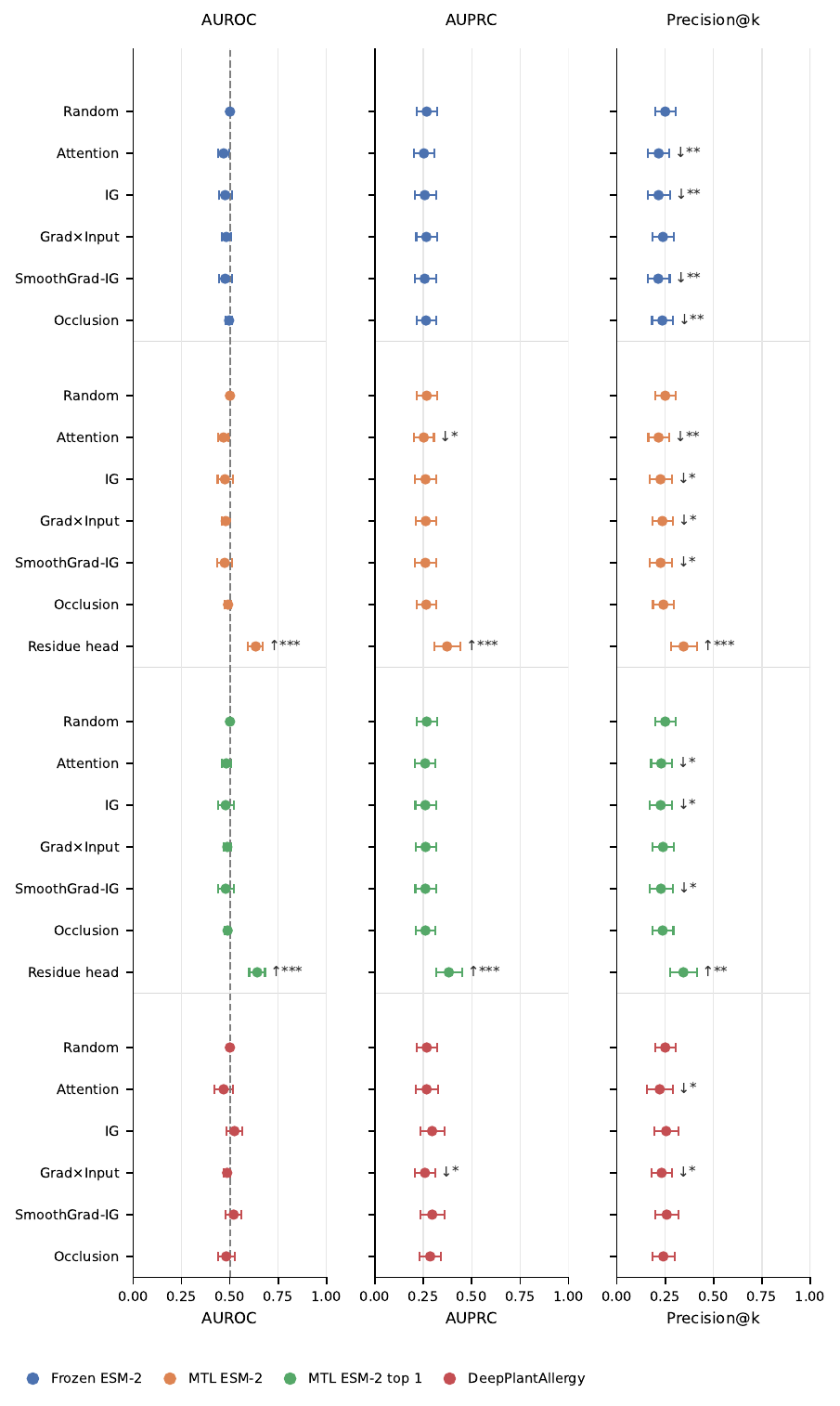}
  \caption{
  Supplementary all-signals faithfulness analysis. Rows are grouped by model
  family and show residue-level AUROC, AUPRC, and Precision@$k$ for each
  available signal. Points show mean performance with bootstrap 95\% confidence
  intervals. Significant paired Wilcoxon comparisons against the corresponding
  within-model random baseline are annotated after Benjamini--Hochberg
  correction; non-significant comparisons are omitted for readability.
  }
  \label{fig:all_signals_significance}
\end{figure}

Results were consistent with \cref{sec:results}; see \cref{fig:all_signals_significance} for the full signal breakdown.

\section{Main Residue-Alignment Summary Table}
\label{app:alignment_table}

\begin{table}[h]
\centering
\caption{
Main residue-level alignment summary on 61 held-out allergenic proteins.
Values are mean per-protein metrics with bootstrap 95\% CIs.
}
\label{tab:alignment_summary}
\tiny
\setlength{\tabcolsep}{2.4pt}
\resizebox{\columnwidth}{!}{%
\begin{tabular}{lcccccc}
\toprule
Signal
& AUROC & 95\% CI
& AUPRC & 95\% CI
& P@$k$ & 95\% CI \\
\midrule
Random
& 0.501 & [0.499, 0.502]
& 0.267 & [0.218, 0.320]
& 0.250 & [0.200, 0.303] \\
Frozen IG
& 0.476 & [0.443, 0.511]
& 0.257 & [0.206, 0.315]
& 0.216 & [0.162, 0.275] \\
Frozen occlusion
& 0.496 & [0.480, 0.511]
& 0.264 & [0.214, 0.315]
& 0.235 & [0.182, 0.292] \\
MTL auxiliary residue head
& 0.634 & [0.594, 0.672]
& 0.372 & [0.307, 0.442]
& 0.345 & [0.282, 0.413] \\
DeepPlantAllergy IG
& 0.524 & [0.480, 0.565]
& 0.295 & [0.236, 0.359]
& 0.255 & [0.194, 0.318] \\
\bottomrule
\end{tabular}%
}
\end{table}

\section{Label-Scrambling Sanity Check}
\label{app:scrambling}

\begin{figure}[h]
  \centering
  \includegraphics[width=\linewidth]{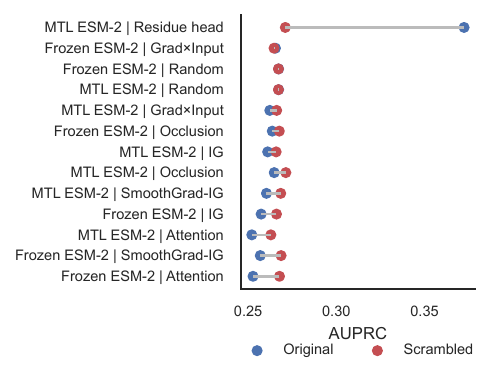}
  \caption{Label-scrambling sanity check. Epitope labels are permuted within
  each protein while residue scores are held fixed ($n=61$). Rows are sorted
  by $\Delta\text{AUPRC} = \text{original} - \text{scrambled}$.}
  \label{fig:label_scrambling}
\end{figure}

To verify that the residue-level evaluation is free of metric bias or
label leakage, we permuted epitope labels within each protein while holding
residue scores fixed. As shown in \cref{fig:label_scrambling}, the MTL
ESM-2 residue head suffered the largest collapse ($0.372 \to 0.271$), while
post-hoc attribution methods remained within $\pm 0.02$ of their original
values. This larger drop is consistent with the residue head learning
epitope-related information during MTL supervision, while scrambled AUPRC
values for other methods clustered near the empirical random-ranking AUPRC
baseline ($\sim0.267$), consistent with random behavior.

\section{Functional Masking Details}
\label{app:masking}

The masking sweep evaluated $k=\{5,10,20,25,30,35,40,45,50\}\%$ on
high-confidence allergenicity predictions with baseline probability above 0.70.
The selected threshold was $k=40\%$, which maximized the fraction of validated
proteins, with ties broken by larger mean $\Delta p$ and then smaller $k$. At
$k=40\%$, 37 of 46 proteins exceeded the validation threshold $\Delta p>0.05$.
IG-vs-random masking yielded a paired Wilcoxon $p=1.83\times 10^{-9}$.

This analysis was conditioned on proteins that the classifier predicted as
allergens with high confidence. It tested whether IG identified residues
causally important for confident positive model predictions, not whether IG
masking has the same effect across all proteins.

As a sensitivity analysis, we repeated the same masking procedure without the
confidence filter, using all 61 splitB allergenic proteins. At the same
operating point $k=40\%$, IG-guided masking remained substantially stronger than
random masking of equal size (IG mean $\Delta p=0.238$ vs.\ random $-0.023$;
paired Wilcoxon $p=3.7\times10^{-7}$), indicating that the effect was not driven
by the confidence-based restriction.

\begin{figure}[h]
  \centering
  \includegraphics[width=\linewidth]{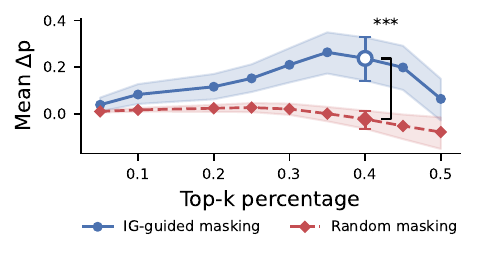}
  \caption{
  Unfiltered sensitivity analysis of IG-guided masking over all 61 splitB
  allergenic proteins. The confidence filter $F(x)>0.70$ is removed; all other
  masking and random-baseline procedures match the main analysis. IG-guided
  masking remains stronger than random masking of equal size at the selected
  operating point $k=40\%$.
  }
  \label{fig:ig_masking_unfiltered}
\end{figure}
\section{Saturation Mutagenesis Details}
\label{app:mutagenesis}

\begin{figure}[h]
  \centering
  \includegraphics[width=\linewidth]{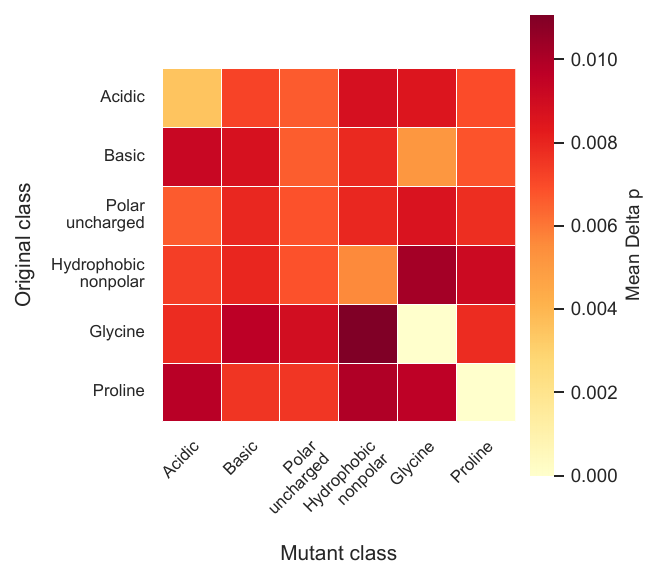}
  \caption{
  Charge/polarity transition heatmap for saturation mutagenesis. Each cell
  reports mean $\Delta p$ for substitutions from an original residue class to
  a mutant residue class.
  }
  \label{fig:charge_polarity_heatmap}
\end{figure}

\begin{figure}[h]
  \centering
  \includegraphics[width=\linewidth]{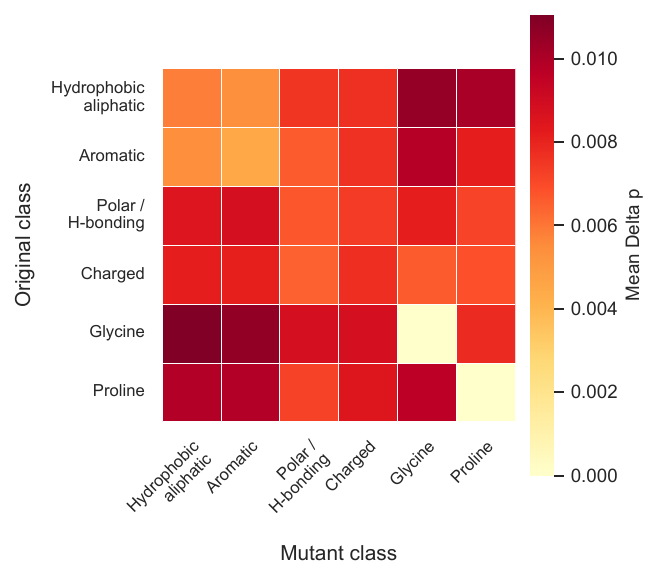}
  \caption{
  Hydrophobicity/aromaticity transition heatmap for saturation mutagenesis.
  Each cell reports mean $\Delta p$ for substitutions from an original
  residue class to a mutant residue class.
  }
  \label{fig:hydrophobicity_heatmap}
\end{figure}

Saturation mutagenesis was run on the 37 proteins validated by IG masking at
$k=40\%$ (\cref{app:masking}). Each residue was replaced with all 19
alternative amino acids. Effects were summarized by original residue identity
(\cref{fig:transition_scatter}) and residue-class transitions
(\cref{fig:charge_polarity_heatmap,fig:hydrophobicity_heatmap}).


\subsection{Residue-Class Definitions}
\label{app:mutagenesis_classes}

We aggregate substitutions using two coarse biochemical groupings. These
categories were not used during training; they were used only to summarize
whether model sensitivity follows interpretable biochemical axes.

\begin{table}[h]
\centering
\caption{
Residue classes used for mutagenesis summaries. Glycine and proline are
listed individually because their effects on backbone geometry are not
reducible to a shared physicochemical class. The remaining classes are
coarse biochemical interpretation aids rather than mutually exclusive
mechanistic claims.
}
\label{tab:residue_classes}
\scriptsize
\setlength{\tabcolsep}{3.0pt}
\resizebox{\columnwidth}{!}{%
\begin{tabular}{lll}
\toprule
Grouping & Class & Residues \\
\midrule
Charge/polarity
& Acidic (negatively charged)  & D, E \\
& Basic (positively charged)   & K, R, H \\
& Polar, uncharged              & S, T, N, Q, Y, C \\
& Hydrophobic, nonpolar         & A, V, L, I, M, F, W \\
& Glycine                       & G \\
& Proline                       & P \\
\midrule
Hydrophobicity/aromaticity
& Hydrophobic, aliphatic        & A, V, L, I, M \\
& Aromatic                      & F, W, Y \\
& Polar or hydrogen-bonding     & S, T, N, Q, C \\
& Charged                       & D, E, K, R, H \\
& Glycine                       & G \\
& Proline                       & P \\
\bottomrule
\end{tabular}%
}
\end{table}

\subsection{Epitope vs.\ Non-Epitope Mutagenesis Sensitivity}
\label{app:epitope_vs_non_epitope}

To directly test whether the model assigns greater functional weight to
epitope residues, we ran full-sequence saturation mutagenesis on all 61
splitB positive proteins and compared mean $|\Delta p|$ averaged over
epitope versus non-epitope residues within each protein. Non-epitope
residues showed marginally but significantly larger sensitivity than
epitope residues (paired Wilcoxon $p=0.021$, two-sided), with no
significant difference in signed effect or fraction of reducing
substitutions ($p>0.87$ for both). This rules out the possibility that
the model assigns greater functional weight to epitope positions and
provides a third independent line of evidence for global rather than
epitope-localized sensitivity.

\begin{figure}[h]
  \centering
  \includegraphics[width=0.72\linewidth]{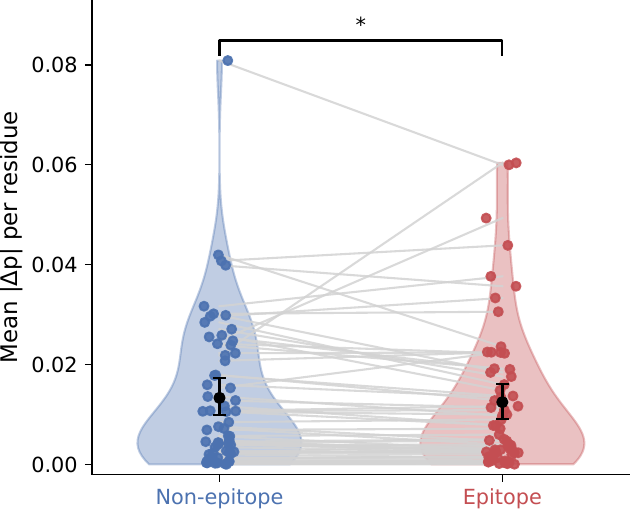}
  \caption{
  Paired comparison of mean $|\Delta p|$ per residue between non-epitope
  and epitope regions across all 61 splitB allergenic proteins. Each
  point is one protein; lines connect paired values. Black markers show
  bootstrap mean $\pm$ 95\% CI. Non-epitope residues show marginally
  larger sensitivity than epitope residues (Wilcoxon $p = 0.021$,
  two-sided; mean paired difference $= -0.0009$,
  95\% CI $[-0.0027, 0.0012]$).
  }
  \label{fig:epitope_vs_non_epitope}
\end{figure}
\end{document}